\documentclass[letterpaper, 10 pt, conference]{ieeeconf}  

\usepackage{multirow}
\usepackage{siunitx}
\usepackage{csvsimple}
\usepackage{graphicx}
\usepackage{caption}
\usepackage{subcaption}
\usepackage{url}

\usepackage{todonotes}

\IEEEoverridecommandlockouts                             

\begin{document}

\title{
Estimation of Kinematic Motion from Dashcam Footage}

\author{Evelyn Zhang$^{1}$, Alex Richardson$^{2}$, and Jonathan Sprinkle$^{3}$
\thanks{$^{1}$Evelyn Zhang from Vanderbilt University,
        Nashville, Tennessee, USA
        {\tt\small evelyn.k.zhang@vanderbilt.edu}}%
\thanks{$^{2}$Alex Richardson from Vanderbilt University,
        Nashville, Tennessee, USA
        {\tt\small william.a.richardson@vanderbilt.edu}}%
\thanks{$^{3}$Jonathan Sprinkle from Vanderbilt University,
        Nashville, Tennessee, USA
        {\tt\small jonathan.sprinkle@vanderbilt.edu}}%
}

\maketitle

\begin{abstract}
The goal of this paper is to explore the accuracy of dashcam footage to predict the actual kinematic motion of a car-like vehicle. Our approach uses ground truth information from the vehicle's on-board data stream, through the controller area network, and a time-synchronized dashboard camera, mounted to a consumer-grade vehicle, for 18 hours of footage and driving. The contributions of the paper include neural network models that allow us to quantify the accuracy of predicting the vehicle speed and yaw, as well as the presence of a lead vehicle, and its relative distance and speed. In addition, the paper describes how other researchers can gather their own data to perform similar experiments, using open-source tools and off-the-shelf technology.

\end{abstract}

\section{Introduction and Related Works}
One application of the extraction of kinematic data from dashcam videos is the ability to detect specific events that occur while driving from just the footage, such as turns, lane changes, and passes. Others have employed different methods for accomplishing this, such as training videos to descriptions of the videos. However, this relies on having existing descriptions of the events in the videos. Some use a combination of GPS/IMU data and footage with marked objects, but again, this relies on having objects in videos marked by hand correctly. Finally, others seek to mark less complex driving events such as a car being stationary, slow, normal, or turning, which are events that can be marked if kinematic data can be derived from footage \cite{9913431,9686618,8569952,10.1145/3383812.3383817}.

In order to train neural networks to produce predictions that are reliable enough to use for other applications, a large dataset is necessary. Currently, BDD100k \cite{yu2020bdd100k} is among the largest available datasets \cite{caesar2020nuscenes} with over a thousand hours of footage. Although there is an ample amount of footage, speed is the only reliable kinematics data available, with regard to the kinematics data important to this study. The BDD100k also had accelerometer data, but due to inconsistent collection of data causing roughly half the values to be negative, the data cannot be trained on.

Our dataset is collected from a single vehicle driven primarily in Arizona. The data include arterial and city roads, freeways, night and day, and multiple types of weather. We collect CAN data directly from the car, which is synchronized with video data from dashcam footage. This allows us to train and predict a more diverse set of kinematics. For this experiment, we use a total of 18 hours of footage, with 15.5 hours acting as our training set, and 2.5 hours used as a validation dataset to verify our results. 

Other datasets exist such as KITTI \cite{KITTI}, ApolloScape \cite{apollo}, Waymo Open \cite{waymo}, etc. However, many are smaller than 18 hours, and others do not contain the data we want to train on, such as different times of day and different types of weather or the various kinematics we use.

The data we have collected is from the controller area network (CAN) bus, which is collected directly from the vehicle. In order to understand the collected CAN messages, we use the Strym tool \cite{bhadani2022strym} to decode the relevant CAN messages into specific kinematic data. This allowed us to extract and convert specific data points into usable tensors for training.

\section{Problem Formulation}
We describe our approach to designing our final experiment, and the motivations behind certain choices. We also describe the preliminary testing performed to collect context for final test parameters and design.

\subsection{Desired Attribute Prediction}
One application that we are hoping to use the results of this study for is the ability to automatically mark specific events in footage, such as turns, overtaking a car, lane switches, etc. Marking these events in footage will help us better understand the contents of the videos and allow us to perform further analysis.

In order to determine these events from just footage, we train the models on speed, yaw, the presence of a lead car, the distance of a lead car, and the relative speed of a lead car. Once these have been predicted by the model, we can automatically find events by designing a set of conditions to be met by the predicted values (e.g., yaw exceeding a certain number indicates a right turn). Additionally, focusing on using models to predict this set of attributes allows us to add events that we have not considered yet without requiring more training, as opposed to directly training on marked events. 

\subsection{BDD100k Preliminary Testing}
Preliminary testing was performed on the BDD100k (Berkeley DeepDrive 100k) dataset in order to ensure that some level of learning was achievable, and to learn what reasonable paramters were for experiments of this type and scale. 

The BDD100k dataset is split into 40 second chunks of video footage collected primarily in New York, San Francisco, and Berkeley, totaling to about 1000 hours. The dataset is comprehensive in comparison to others previously mentioned, containing footage taken in the rain, at night, during cloudy days, and on highways. For our testing, we primarily focused on the speed and yaw data, which was collected using phones positioned on the dashboard of the cars.

The dataset needed to be modified in order to be used for our training purposes. The videos were stretched to form a square and the videos were set to 5 frames per second. With too many frames per second, training takes too long, but with not enough, information is lost. We leave to future work the exploration of frame rate as a hyperparameter.

The first round of testing we performed was using the U-NET autoencoder \cite{unetautoencoder}. The BDD100k dataset includes one frame per video which contains annotations of objects, and the autoencoder was trained on these frames. Every video was processed into 512 latents using this autoencoder, and a simple grid search was conducted for speed. There were four models: one non-recurrent baseline neural network, two recurrent neural networks of GRU \cite{chung2014empirical} and LSTM \cite{lstm}, and a transformer \cite{attentionisallyouneed}. 
The results of our testing are shown in Table~\ref{tab:accuracy_bdd}. This shows the results of three learning rates, with each model running for 50 epochs, and a batch size of 32, two layers in the neural network, and the RNNs used a hidden size of 512.

\renewcommand{\arraystretch}{1.3}
\begin{table}[htbp]
    \centering
    \begin{tabular}{ |c||c|c|c|c|  } 
    \hline
    LR & Baseline & GRU & LSTM & Transformer\\
    \hline\hline
    \num{1e-1} & 25.64 & 52.32 & 50.57 & 44.96 \\ 
    \num{1e-3} & 19.77 & 8.97 & 10.07 & 14.63 \\ 
    \num{1e-5} & 44.23 & 12.88 & 14.23 & 15.36 \\
    \hline
    \end{tabular}
    \caption{Validation Set MSE (m/s) of Models after 50 epochs}
    \label{tab:accuracy_bdd}
\end{table}
The use of an autoencoder can be effective if the output targets can be fine tuned for ground truth information. However, in the context of this work, the use of an autoencoder will have limitations, which motivates the use of a convolutional neural network (CNN) which allows us to retrain the algorithms on the BDD dataset, to establish a baseline for comparison against our own dataset in the future sections.
The best results of our CNN had the setup of [3, 32, 64, 64, 128, 256] with convolution blocks of [1,1,2,2,4,4] (14 layers total). 
To combat the vanishing gradient issue 
we utilized a residual neural network design, with an
identity function was applied at the end of each convolution block. The CNN structure resulted in a 1D tensor of 512, which matches the autoencoder output target for ease of comparison of results.

Our best model was able to reach a mean squared error of 2. This model had a batch size of 32 and a learning rate of 0.0025. Visually inspecting the footage, the predicted speed stayed relatively close to the actual speed. This demonstrated to us that the results we were looking for are obtainable. 
Inconsistency in kinematic data collection in the BDD100k dataset resulted in difficulty training the models for yaw, so we were unable to test yaw on the BDD100k dataset.

\subsection{Video and CAN Dataset}
The dataset we used is comprised of 18 hours of data, formatted as a set of separate videos, and csv files, containing the dashcam footage and CAN data respectively. The videos and data needed to be processed into something that could be trained on. Firstly, they were all synchronized, so that the videos matched the data at any given time. The usable blocks of time that had both types of data were then extracted to be used (some timestamps would have just footage and no CAN data and vice versa). 

Using Strym, we parsed the csv files and extracted just the attributes we planned to use (speed, yaw, lead present, lead distance, lead speed) into json files, and named them accordingly. These attributes were also linearly interpolated into 5 frames per second and cut to 40 seconds per video.

The final dataset for this experiment amounted to about 18 hours, with 15.5 hours used to train the models, and 2.5 models to validate the training. This is not much in comparison to the BDD100k dataset, so results will not have a comparable MSE.

\section{Model Structure and Training}
\begin{figure}[htbp]
    \centering
    \begin{subfigure}[t]{0.5\columnwidth}
        \includegraphics[width=\linewidth]{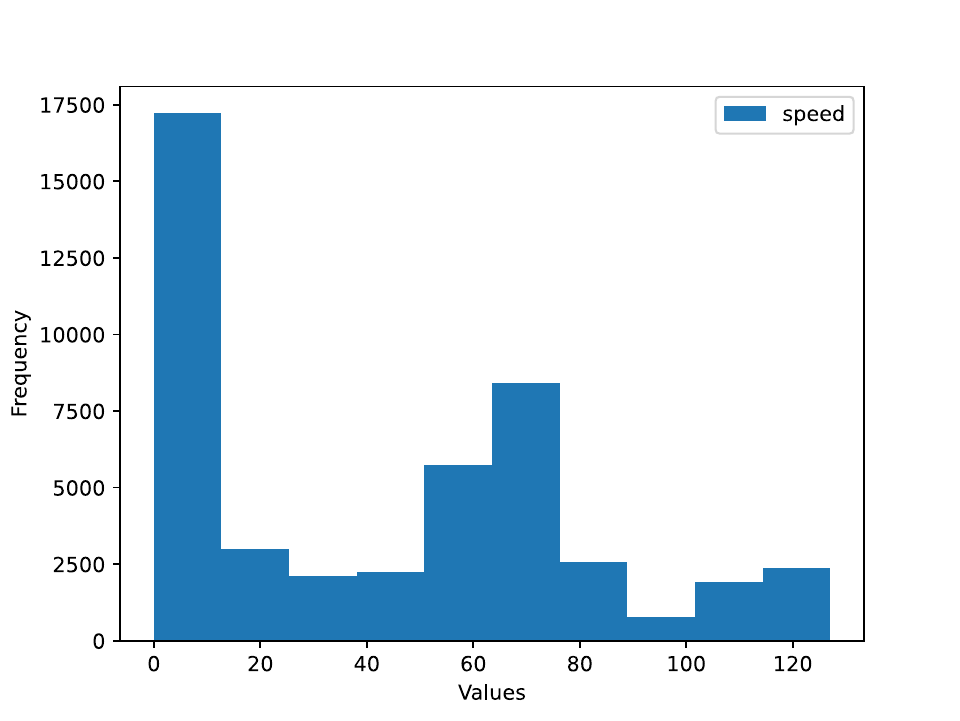}
        \caption{Speed (km/h) vs Frequency}
        \label{fig:speed_a}
    \end{subfigure}\hfill
    \begin{subfigure}[t]{0.5\columnwidth}
        \includegraphics[width=\linewidth]{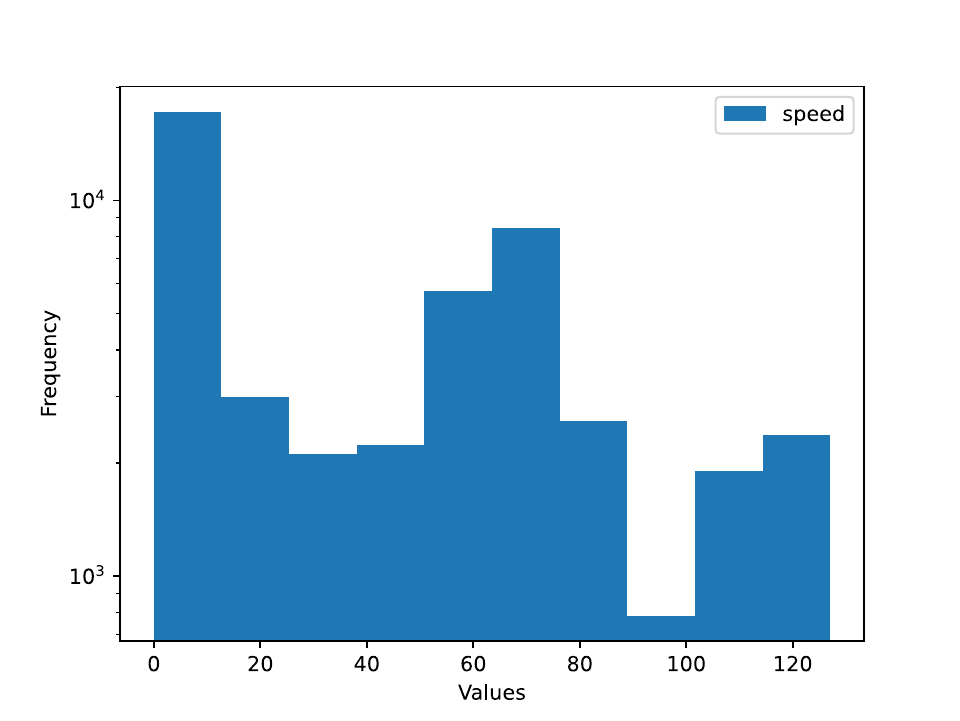}
        \caption{Speed (km/h) vs Frequency Logarithmic Scale}
        \label{fig:speed_b}
    \end{subfigure}
    \begin{subfigure}[b]{0.5\columnwidth}
        \includegraphics[width=\linewidth]{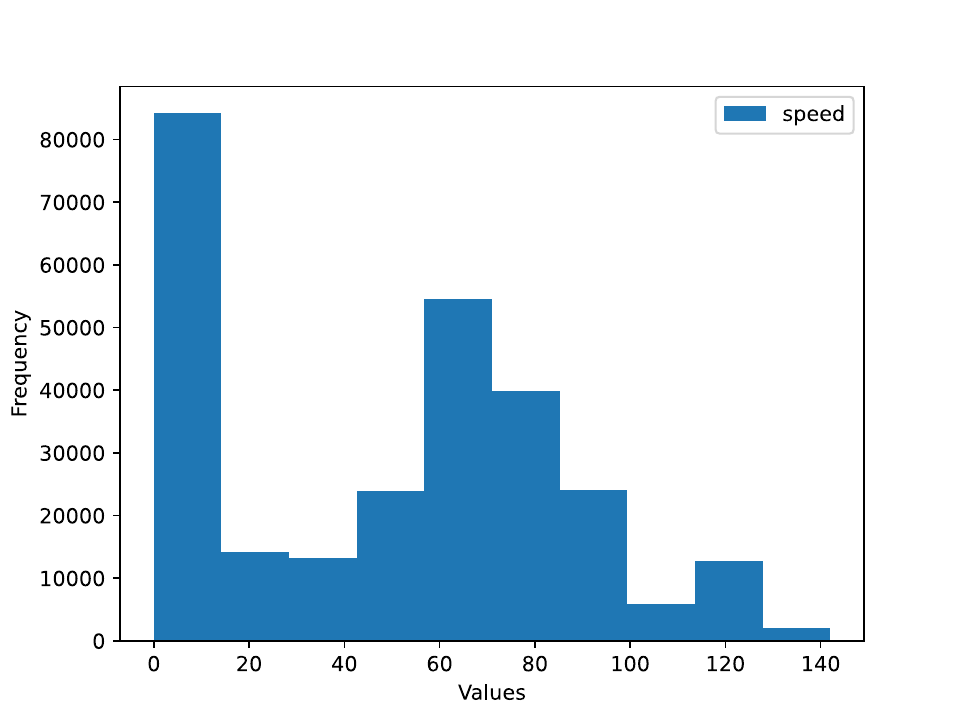}
        \caption{Training Data Speed (km/h) \protect\newline vs Frequency}
        \label{fig:speed_c}
    \end{subfigure}\hfill
    \begin{subfigure}[b]{0.5\columnwidth}
        \includegraphics[width=\linewidth]{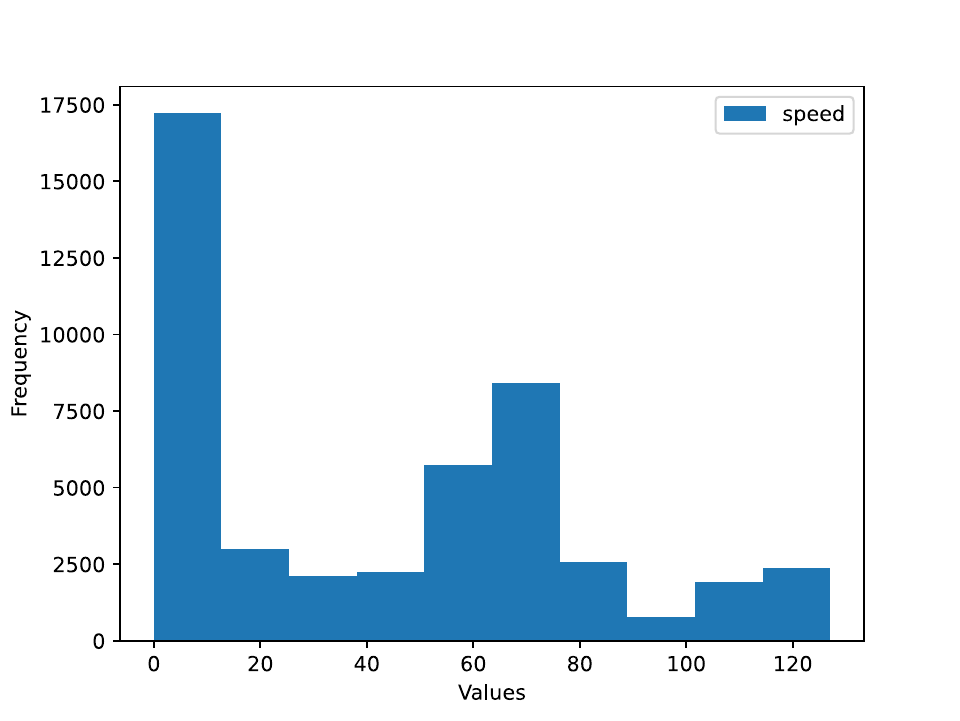}
        \caption{Validation Data Speed (km/h) vs Frequency}
        \label{fig:speed_d}
    \end{subfigure}
    \caption{Graphs displaying spread of data of speed (km/h)}
\end{figure}

Speed is one of the key statistics we are training our models on, for good reason. Nearly every action a car can take can be related to speed in some way. We measure speed in km/h.

Looking at the histogram a), the dataset contains many speeds, with a good number of data points spread across all the values. Because of this, speed will likely have comparatively good results since there exists good trainable data. Additionally, from histograms c) and d), we see that the data in the training and validation sets are somewhat evenly distributed so there is unlikely to be training bias. 

\begin{figure}[htbp]
    \centering
    \begin{subfigure}[t]{0.5\columnwidth}
        \includegraphics[width=\linewidth]{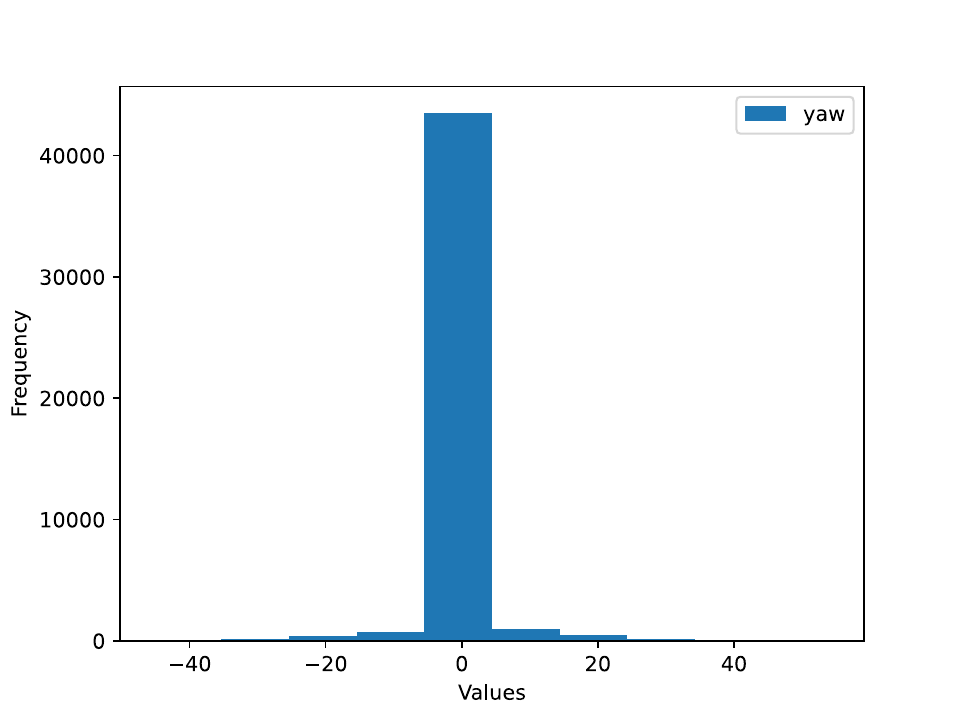}
        \caption{Yaw (deg/s) vs Frequency}
        \label{fig:yaw_a}
    \end{subfigure}\hfill
    \begin{subfigure}[t]{0.5\columnwidth}
        \includegraphics[width=\linewidth]{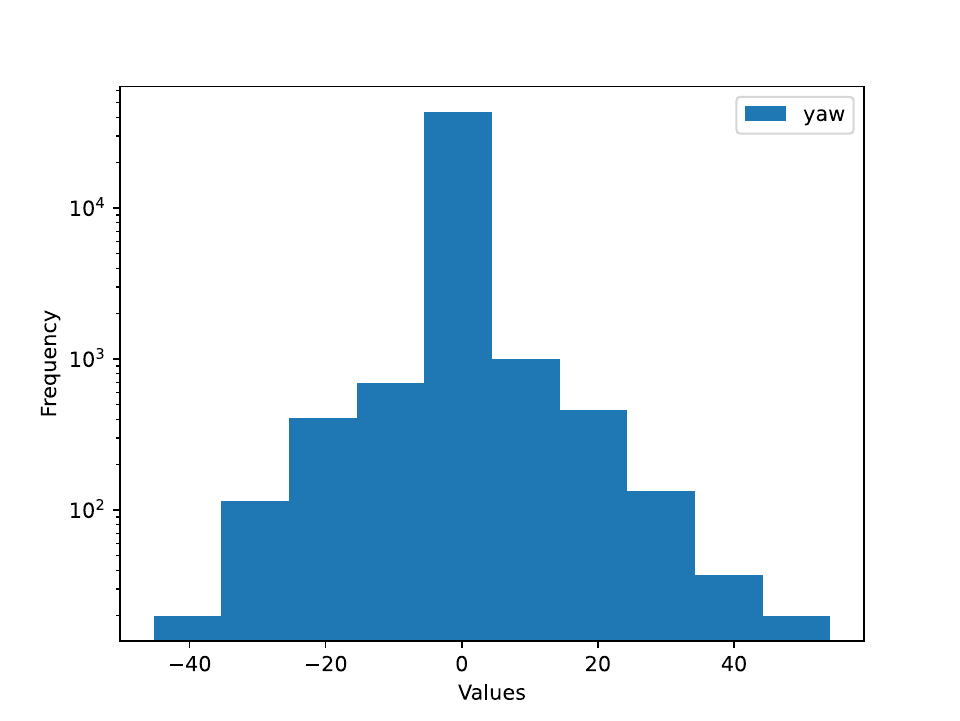}
        \caption{Yaw (deg/s) vs Frequency Logarithmic Scale}
        \label{fig:yaw_b}
    \end{subfigure}
    \begin{subfigure}[b]{0.5\columnwidth}
        \includegraphics[width=\linewidth]{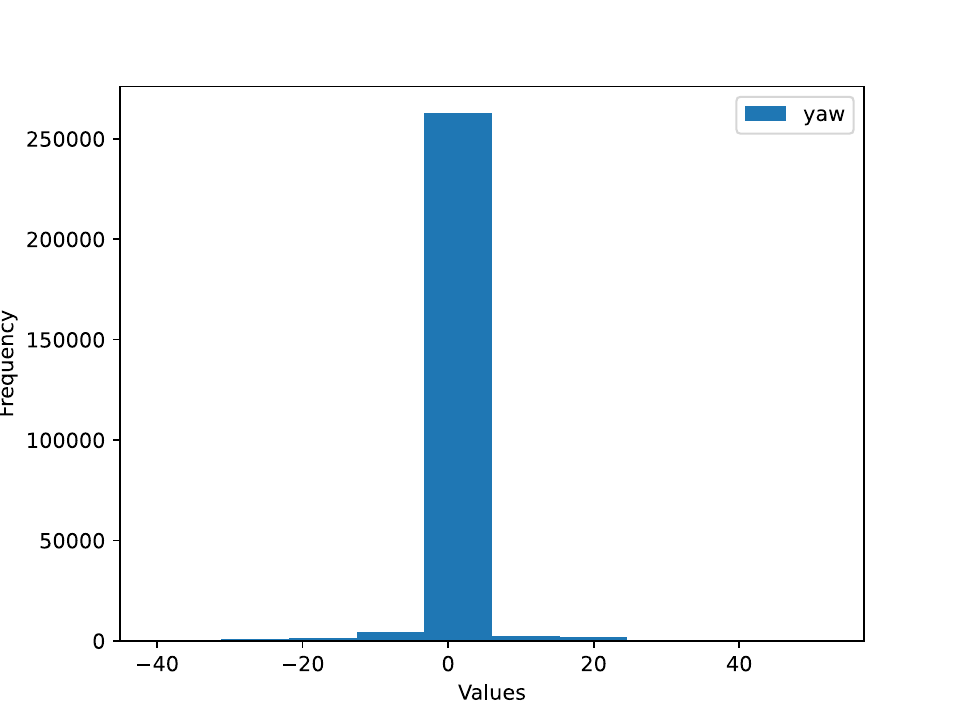}
        \caption{Training Data Yaw (deg/s) \protect\newline vs Frequency}
        \label{fig:yaw_c}
    \end{subfigure}\hfill
    \begin{subfigure}[b]{0.5\columnwidth}
        \includegraphics[width=\linewidth]{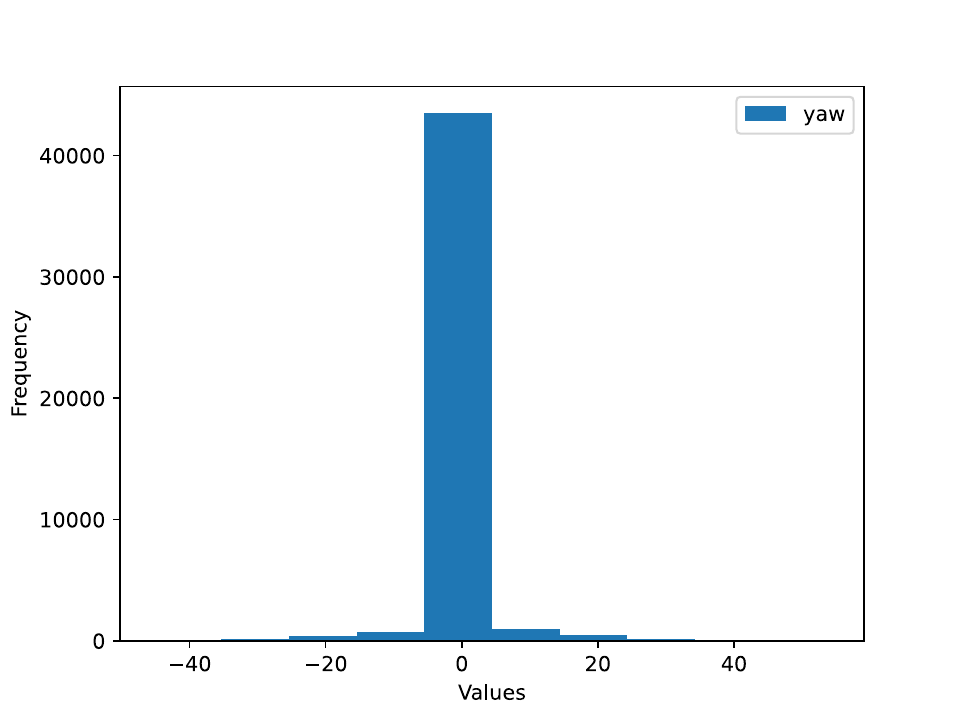}
        \caption{Validation Data Yaw (deg/s) vs Frequency}
        \label{fig:yaw_d}
    \end{subfigure}
    \caption{Graphs displaying spread of data of yaw (deg/s)}
\end{figure}

Next is yaw, vital information in determining turns. A larger magnitude means a larger turn, while negative values indicate left turns and positive values indicate right turns. From the histogram, upwards of 250k points lie between -5 and 5, very close to 0. This indicates that most of our footage contains cars driving relatively straight, and not many turns are in the dataset. This will likely make it more difficult to obtain good results, since there is not that much trainable data on turns. 

The training and validation sets also appear to be relatively similar, so there will be minimal bias as a result of partitioning the dataset.

\begin{figure}[htbp]
    \centering
    \begin{subfigure}[t]{0.5\columnwidth}
        \includegraphics[width=\linewidth]{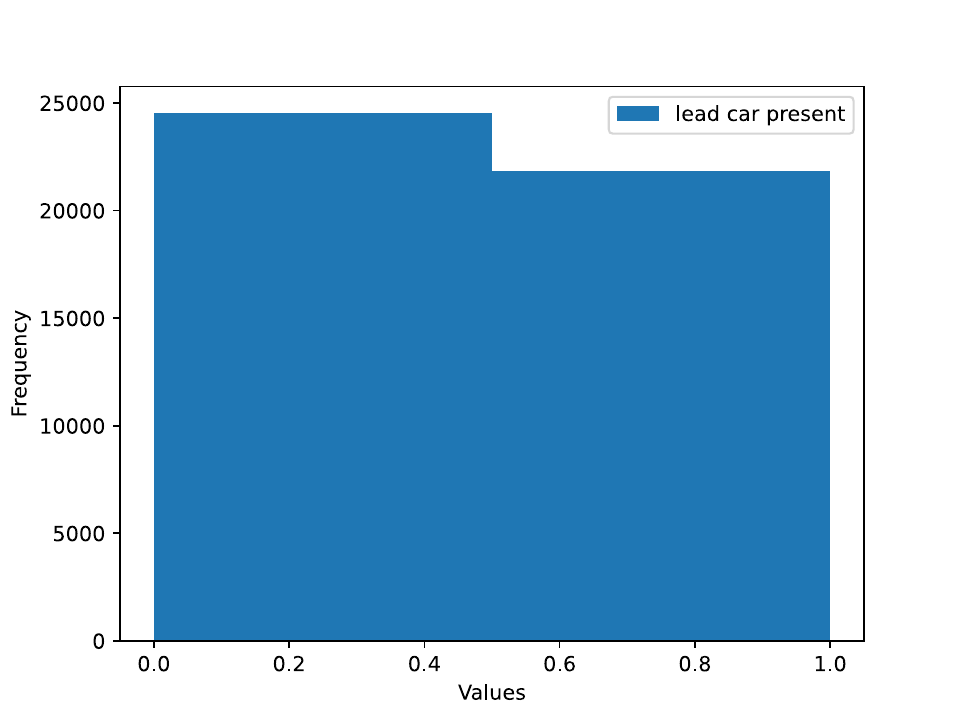}
        \caption{Lead Car Present vs \protect\newline Frequency}
        \label{fig:lcp_a}
    \end{subfigure}\hfill
    \begin{subfigure}[t]{0.5\columnwidth}
        \includegraphics[width=\linewidth]{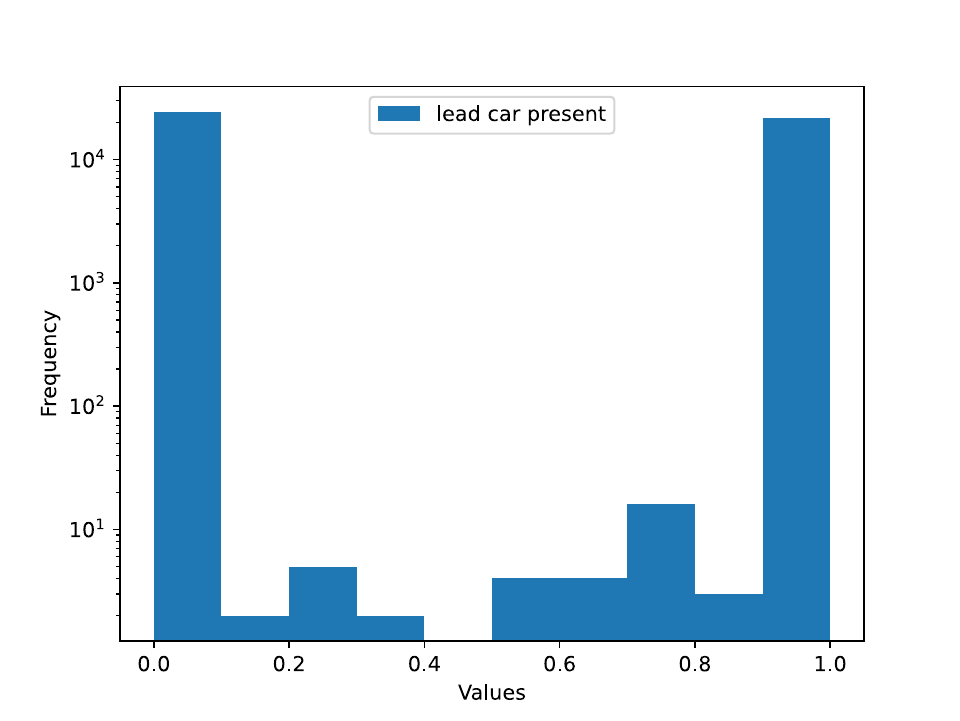}
        \caption{Lead Car Present vs Frequency Logarithmic Scale}
        \label{fig:lcp_b}
    \end{subfigure}
    \begin{subfigure}[b]{0.5\columnwidth}
        \includegraphics[width=\linewidth]{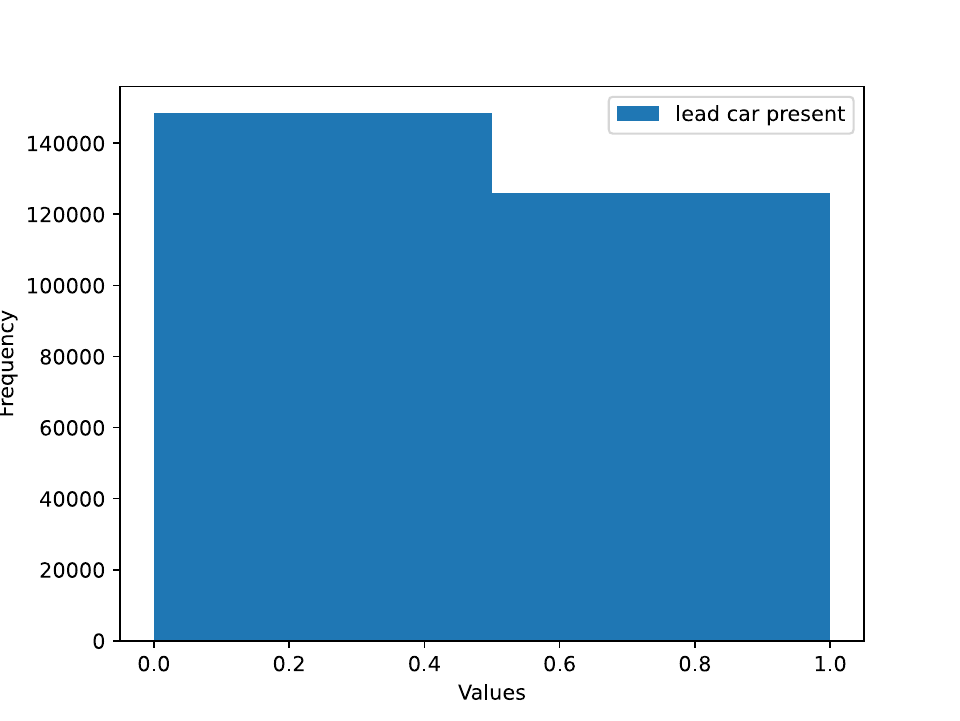}
        \caption{Training Data Lead Car \protect\newline Present vs Frequency}
        \label{fig:lcp_c}
    \end{subfigure}\hfill
    \begin{subfigure}[b]{0.5\columnwidth}
        \includegraphics[width=\linewidth]{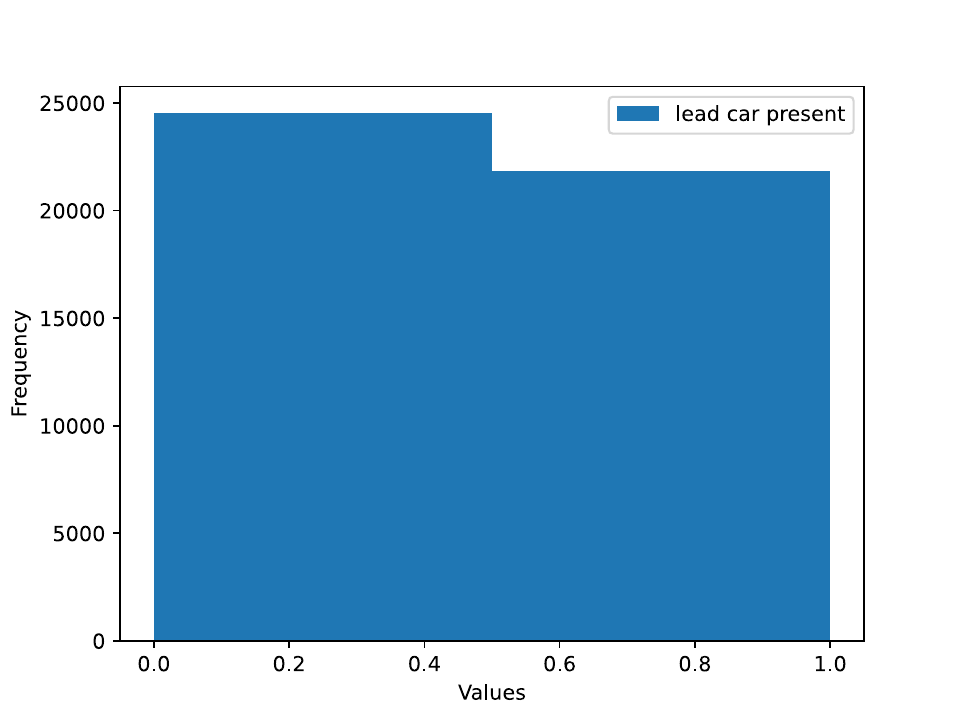}
        \caption{Validation Data Lead Car Present vs Frequency}
        \label{fig:lcp_d}
    \end{subfigure}
    \caption{Graphs displaying spread of data of Lead Car Present, with 1 indicating present and 0 otherwise. Other values are the result of interpolation.}
\end{figure}

The presence of a lead car (a car driving in front of the car with the dashcam) is most important in determining lane switches and overtaking cars. The last two attributes, the distance of the lead car and the relative speed of it, supplement helpful information. It is worth noting that there is a relatively even split between when a lead car is detected and a lead car is not, so strange distributions in the dataset should not be of concern to training. The logarithmic scale shows the presence of values that are the result of interpolation. Again, the distributions between the training data and the validation data are close enough to not be of concern.

\begin{figure}[htbp]
    \centering
    \begin{subfigure}[t]{0.5\columnwidth}
        \includegraphics[width=\linewidth]{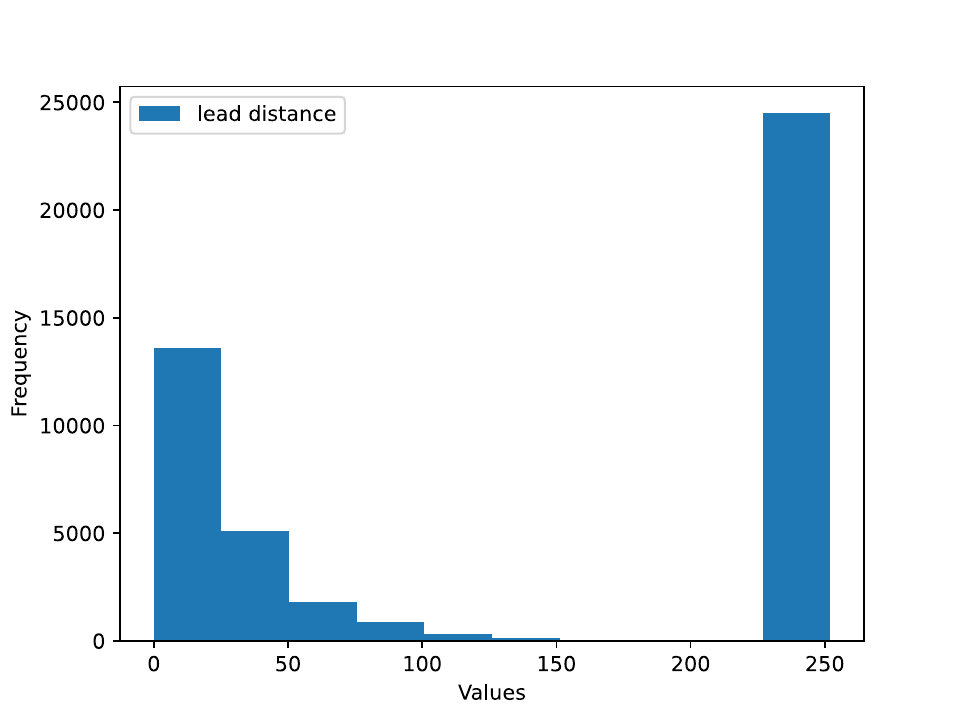}
        \caption{Lead Distance (m) vs \protect\newline Frequency}
        \label{fig:ld_a}
    \end{subfigure}\hfill
    \begin{subfigure}[t]{0.5\columnwidth}
        \includegraphics[width=\linewidth]{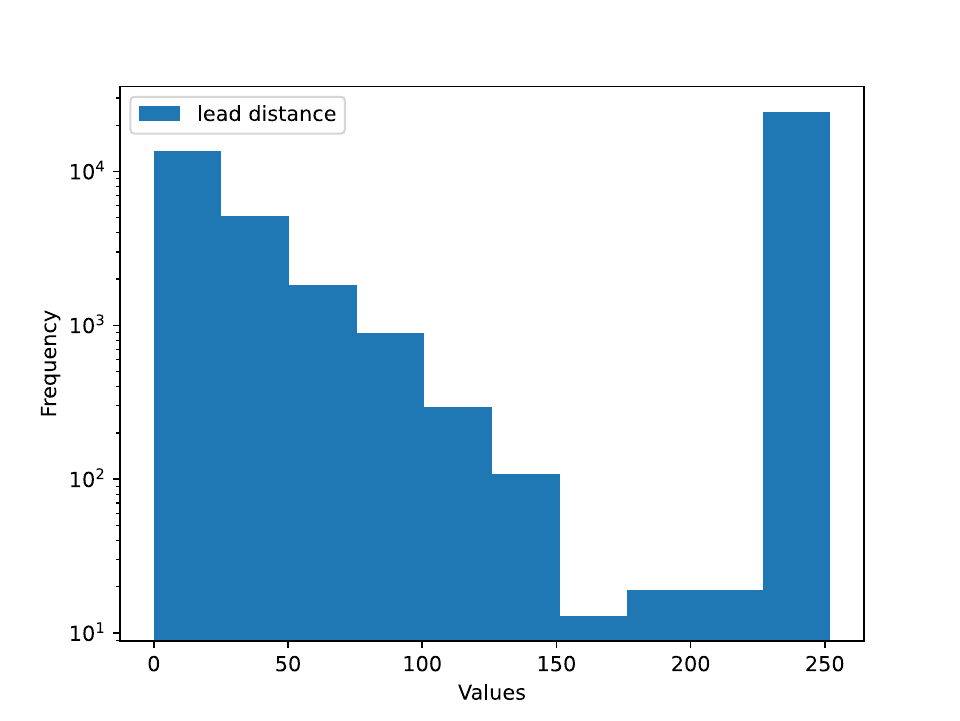}
        \caption{Lead Distance (m) vs \protect\newline Frequency Logarithmic Scale}
        \label{fig:ld_b}
    \end{subfigure}
    \begin{subfigure}[b]{0.5\columnwidth}
        \includegraphics[width=\linewidth]{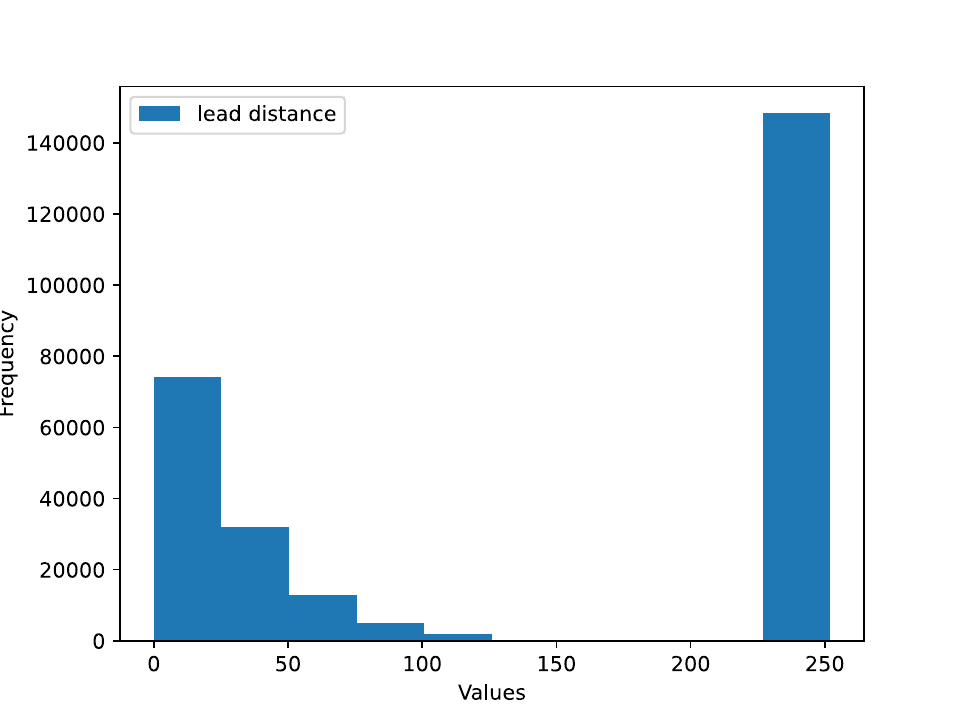}
        \caption{Training Data Lead \protect\newline Distance (m) vs Frequency}
        \label{fig:ld_c}
    \end{subfigure}\hfill
    \begin{subfigure}[b]{0.5\columnwidth}
        \includegraphics[width=\linewidth]{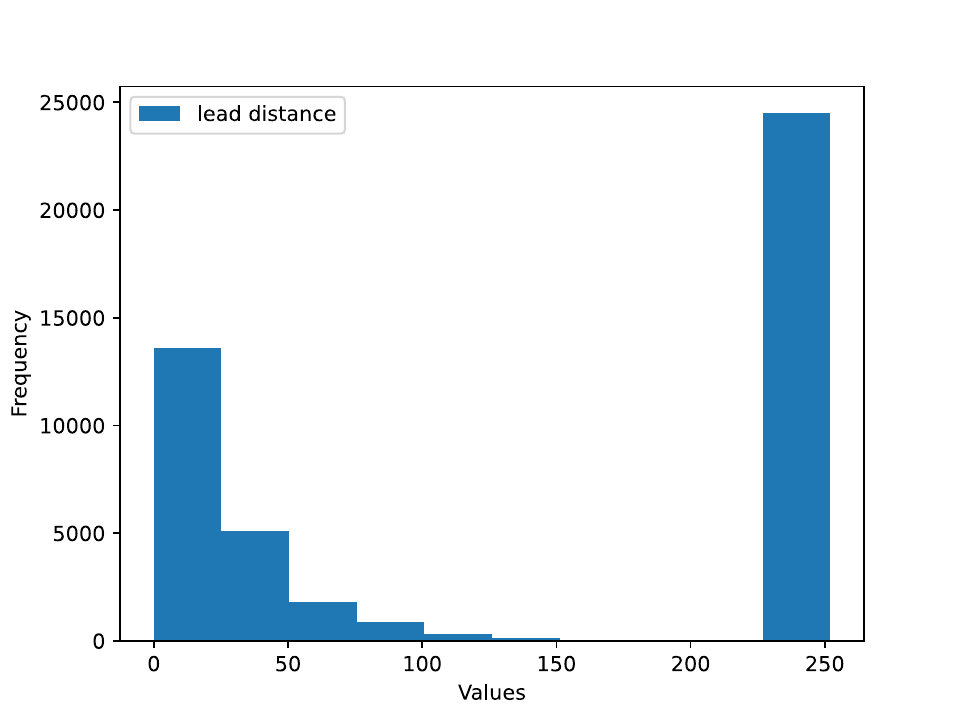}
        \caption{Validation Data Lead Distance (m) vs Frequency}
        \label{fig:ld_4}
    \end{subfigure}
    \caption{Graphs displaying spread of data of Lead Distance (m)}
\end{figure}

However, these distributions could be somewhat concerning for LD (lead distance) and LS (lead speed). For LD, whenever no car is detected, the CAN message says that the distance is 252. This will certainly impact the training quality of LD, not to mention how there is effectively half the amount of training footage, since only half of it has a lead car. The LD distribution is also similar between training and validation sets.

\begin{figure}[htbp]
    \centering
    \begin{subfigure}[t]{0.5\columnwidth}
        \includegraphics[width=\linewidth]{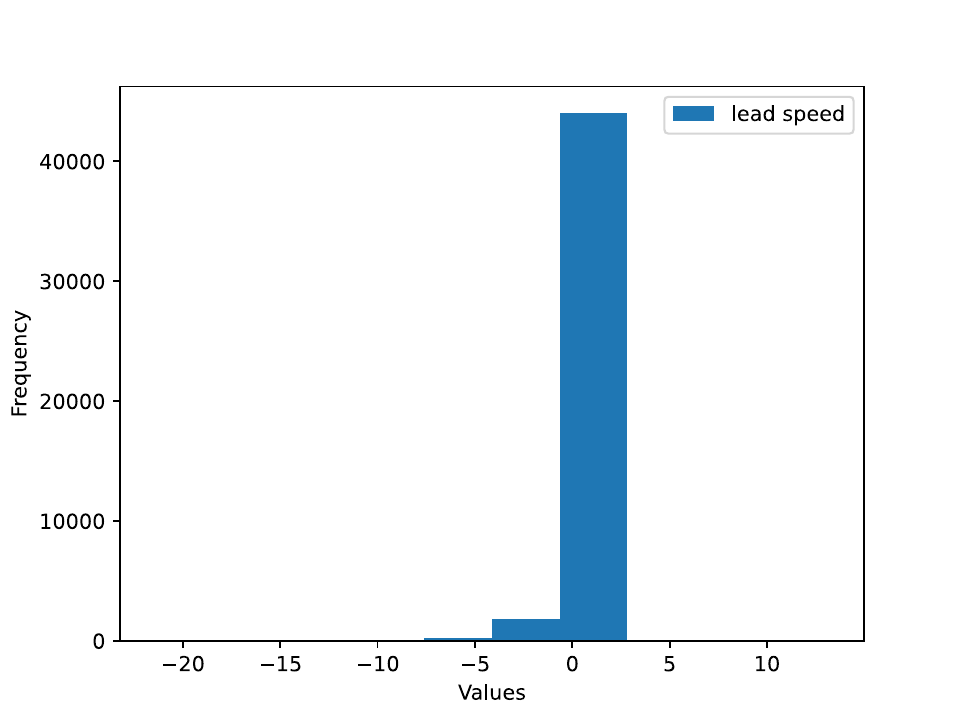}
        \caption{Lead Speed (km/h) vs \protect\newline Frequency}
        \label{fig:ls_a}
    \end{subfigure}\hfill
    \begin{subfigure}[t]{0.5\columnwidth}
        \includegraphics[width=\linewidth]{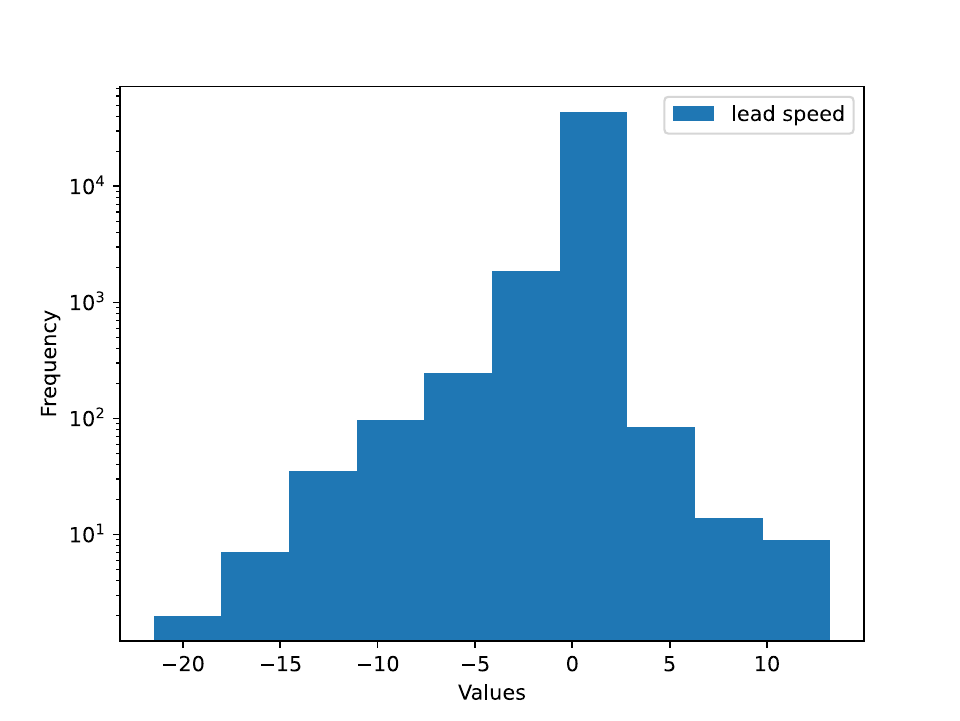}
        \caption{Lead Speed (km/h) vs \protect\newline Frequency Logarithmic Scale}
        \label{fig:ls_b}
    \end{subfigure}
    \begin{subfigure}[b]{0.5\columnwidth}
        \includegraphics[width=\linewidth]{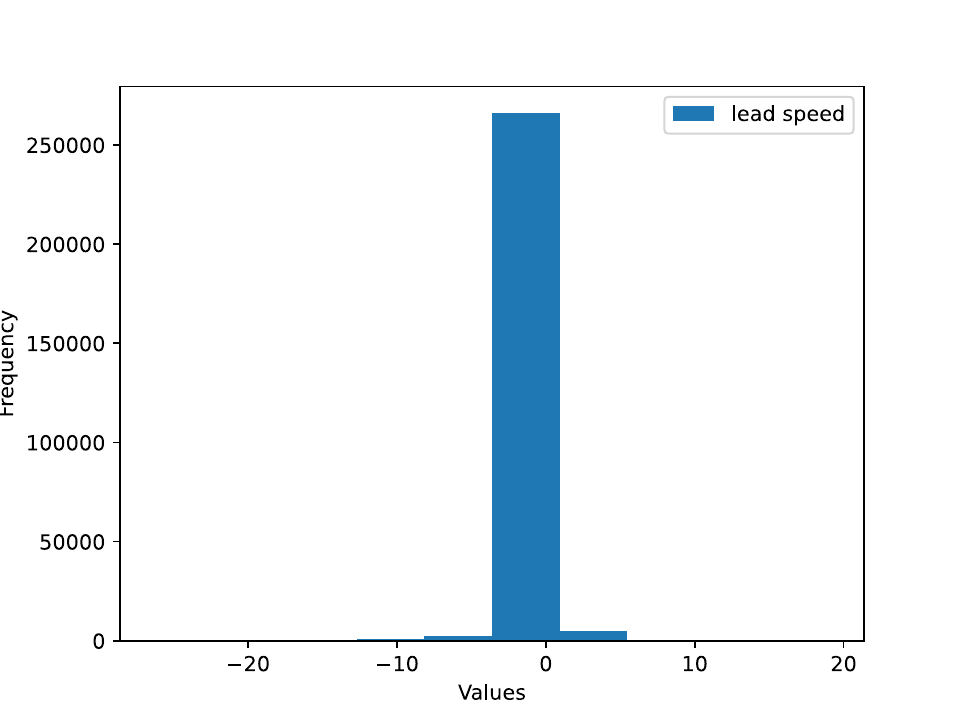}
        \caption{Training Data Lead \protect\newline Speed (km/h) vs Frequency}
        \label{fig:ls_c}
    \end{subfigure}\hfill
    \begin{subfigure}[b]{0.5\columnwidth}
        \includegraphics[width=\linewidth]{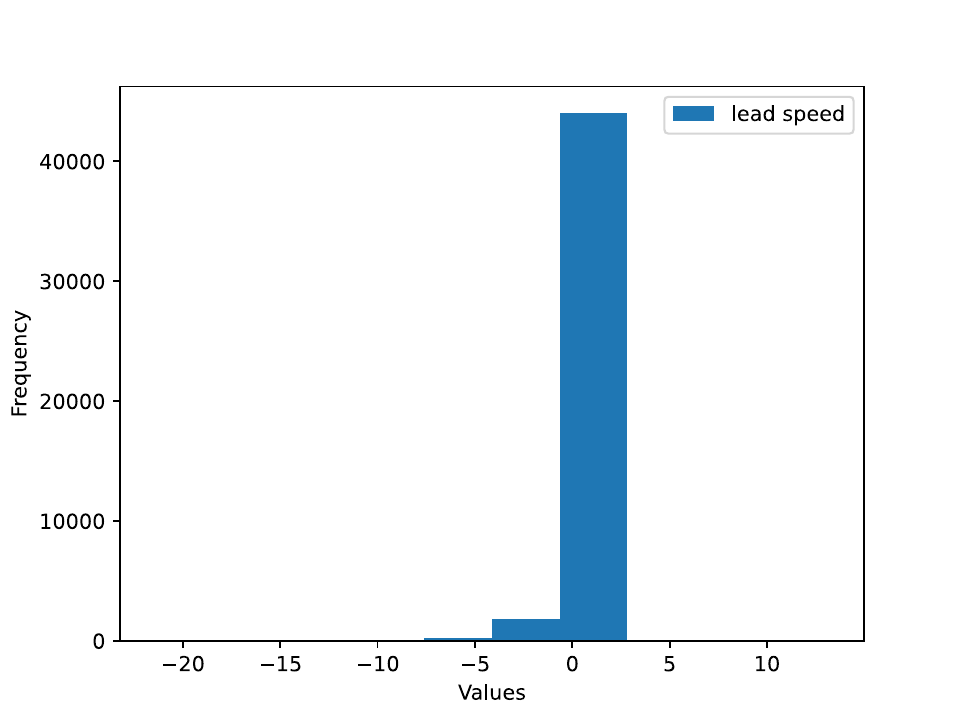}
        \caption{Validation Data Lead Speed (km/h) vs Frequency}
        \label{fig:ls_d}
    \end{subfigure}
    \caption{Graphs displaying spread of data of Lead Speed (km/h)}
\end{figure}

The same goes for LS, since no lead car defaults to a LS of 0. Additionally, when a driver is following another driver, a constant speed is roughly maintained, resulting in many data points staying around 0 regardless. This causes a large cut in useful training data. Unfortunately for LS, there is a noticeable difference between the training data and validation data distributions. The training data is symmetrical while the validation data is noticeably skewed left. Since the training data is symmetric, bias is may not be a concern, but the vast amount of training data close to 0 may mean that there is not much to train on. 

Finally, discrepancy between ground truth and what a human may consider a "lead car" should be mentioned. Any information regarding a lead car is collected using radar, so objects that are not lead cars could be detected, or lead cars may not be detected at times. The effectiveness of detecting a lead car may also be questionable at distances. It is speculated that the strange dip in the data points in the LD histogram after 125 is attributed to this.

\subsection{Model Limitations}
Some important limitations needed to be kept in mind while designing the experiment. The two most important limitations were the amount of time we had and GPU space. 

In regards to the amount of time, we had to limit the number of traits our grid search included, or else the experiment could have required weeks to run. Earlier in the development process, we preprocessed the videos. This cut down on runtime, so it allowed us more flexibility with experiment size. We applied a similar process to other parts of the experiment (the autoencoder, which is mentioned in more detail later) to shorten run time, but limiting the experiment scope was our best method. This meant that we could not test every model and variable we wanted to test, and that we had to limit the number of epochs each model could train for.

In terms of GPU space, we also used preprocessing to limit the amount of GPU space that was actually required, but we also tested out nonconventional methods of batching (such as distributing smaller batches across several nodes for a larger effective batch size), although we encountered complications that could not be fixed within a reasonable amount of time, so we settled on limiting batch size.

\subsection{Model Information}
Our models had five different traits that we varied throughout training. We had two methods of compressing the images to a 512 latent space, which were the autoencoder and convolutional neural networks. After we reached the 512 latent space stage, the tensors were processed by either a non-recurrent baseline neural network, a GRU model, and a transformer. We varied batch size between 1 and 2, and learning rate between $1x10^3$ and $1x10^5$. For certain traits, we augmented the data in hope that the increased data would reduce overfit and produce better results. Every model ran for 250 epochs. This totals to 120 models that we constructed and tested.

For each neural network, we combined a method of turning the 256x256 images with channel sizes of 3 into 1D tensors of size 512, then trained neural networks on the 512 latent space. 

\subsubsection{Autoencoder}
The first method of compressing the images of 1D tensors is the autoencoder used in preliminary testing, that was trained on frames with marked objects. 

\subsubsection{Convolutional Neural Networks}
The second method was the CNN that was also used previously (in-channels of [3,32,64,64,128,256], blocks of [1,1,2,2,4,4] with a ResNet structure everyblock) since both the autoencoder and the CNN had yielded good results. Now that a more comprehensive grid search was going to be completed, we were in a better position to formally compare how well both did. The autoencoder has the benefit of shielding against overfit, since it was pretrained on other data. This is especially important because overfit is a major concern in a training dataset of only 15.5 hours. 

Since the CNN is trained just on our data, it has the potential to be more sensitive to information in the videos that is valuable for our specific attributes, and it can learn information about the images that was overlooked by the autoencoder. 

\subsubsection{Baseline Model}
Something that sets apart neural network learning on videos as opposed to images is the relationship between each frame of the video. In order to test how learning information throughout the video is compared to just image by image, we use a baseline non-recurrent neural network as a standard of comparison for how well the models that have the ability to learn temporal relationships perform.

\subsubsection{Gated Recurrent Unit}
With this in mind, a recurrent neural network is a good choice (ex: knowing the speed a frame before helps you learn the speed now). While training some models to learn speed on the BDD100k dataset, we learned that GRU and LSTM had roughly the same performance, although GRU typically performed slightly better, so we decided to use only GRU in order to limit experiment time without losing much in terms of results. The GRU model updates a hidden state while training, to keep track of information to be remembered from previous frames.

\subsubsection{Transformer}
In addition to the GRU, we also test the effectiveness of a transformer model in learning temporal relationships. This model examines all 200 frames at once and attempts to learn the importance of each of the other frames in reference to the frame it is currently examining. This also has the potential to learn temporal relationships, albeit in a different way than the GRU model because it has the potential to learn more complex relationships than simply remembering the previous frames.

\subsubsection{Batch Size}
Earlier in the design stage, we experimented with larger batch sizes up to 24, though due to issues with GPU space we discovered that we could only reasonably have batch sizes of up to 1 and 2. However, it is worth noting that the batch sizes are functionally double (2 and 4) because we have two workers per node.

\subsubsection{Learning Rate}
For each of our models, we test them with a learning rate of 1x10-3 and 1x10-5. These rates are chosen because they are somewhat standard learning rates, and have proven to have good results when working on the BDD100k dataset. 

\subsubsection{Augmentation}
Because our dataset size is small, especially in comparison with the BDD100k dataset, we decided to explore options for augmentation to artificially increase the amount of data that we have. This would ideally help us reduce overfit and obtain better results. The options that we considered were flipping the video frames horizontally (preserving all the labels except yaw, which is multiplied by negative one to account for the directional change), and reversing the videos (preserving only lead vehicle present and lead vehicle distance). We also compared these against no augmentation at all, and combined the two. 

After running a couple of models, the horizontal augmentation alone would perform better than no augmentation at all. Models training with any reverse augmentation (alone or with horizontal), would not converge. Although the original experiment design included augmentation in the grid search, consistent better performance of the horizontal augmentation prompted us to settle with just horizontal augmentation, conveniently limiting the amount of testing required.

\subsubsection{Epochs}
Our models train for 250 epochs. Based on previous testing, this number of epochs is estimated to be enough for the models to converge. It’s worth mentioning that the models that use autoencoders generally require less epochs to converge than the convolutional neural networks; however, for the sake of standardization (and testing limitations), we keep each model running for the same number of epochs.

\section{Results}

\subsection{Speed}

\begin{figure*}
  \begin{minipage}[c]{.4\linewidth}
    \centering
    \includegraphics[width=0.9\linewidth]{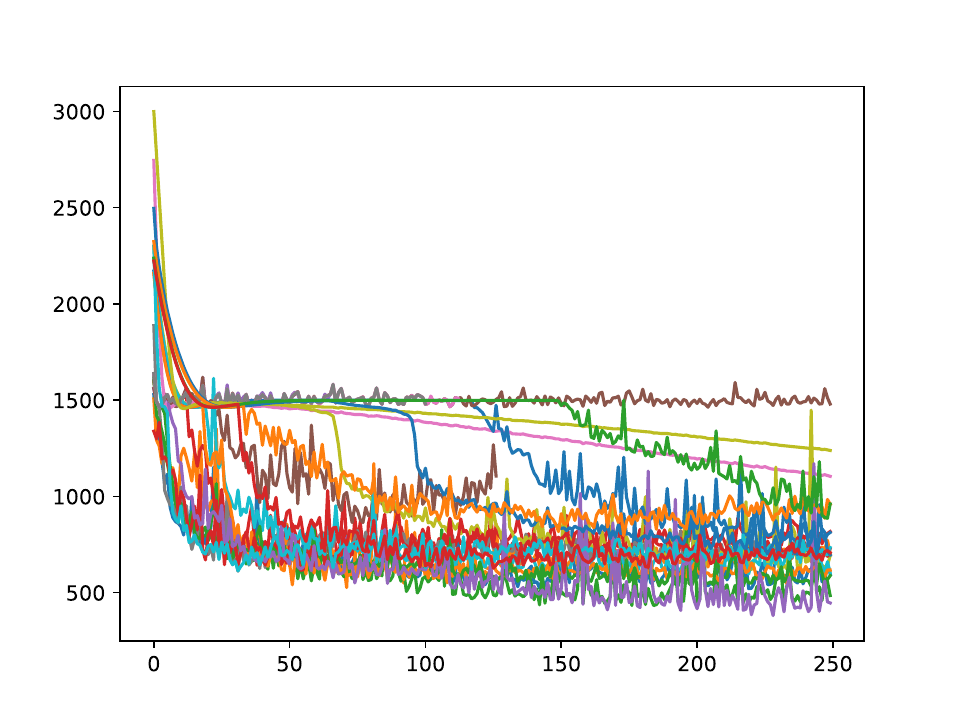}
    \captionof{figure}{Speed Validation Loss (MSE km/h)}
    \label{fig:speed_graph}
  \end{minipage}\hfill
  \begin{minipage}[c]{.6\linewidth}
    \centering
    \begin{tabular}{|c|c|c|c|c|c|c|c|}
        \cline{3-8}
        \multicolumn{1}{c}{} && \multicolumn{2}{|c|}{Baseline} & \multicolumn{2}{|c|}{GRU}&\multicolumn{2}{|c|}{Transformer}\\
        \hline
        BS & LR & CNN & UNET & CNN & UNET & CNN & UNET \\
        \hline
        \multirow{2}{*}{1} & $10^{-3}$ & 720 & 608 & 801 & 522 & NaN & NaN \\
        \cline{2-8}
         & $10^{-5}$ & 703 & 1108 & 630 & 675 & 612 & 810 \\
         \hline
        \multirow{2}{*}{2} & $10^{-3}$ & 728 & 588 & 1479 & 446 & NaN & NaN \\
        \cline{2-8}
         & $10^{-5}$ & 733 & 1238 & 959 & 695 & 707 & 960 \\
        \hline
    \end{tabular}
    \captionof{table}{Speed (MSE km/h)}
    \label{tab:speed_results}
  \end{minipage}
\end{figure*}

The results of the 24 speed models are shown in Table~\ref{tab:speed_results}, and the validation loss across the epochs is displayed in Figure~\ref{fig:speed_graph}. 

Out of all the models, the one using the GRU with an autoencoder learning at a rate of $10^{-3}$ with a batch size of 1 performed the best, resulting in an MSE of 446.253. Many of the models performed similarly well with MSEs in the 500-600 range. This is the only set of neural networks where we can directly compare the results to the BDD100k results, although BDD100k is in m/s while this is in km/hr. When converting units, we have an MSE of 34.433. While this is significantly worse than the results we obtained on the BDD100k dataset, the main difference is the amount of trainable data we have. However, should there be more data, this will be less of an issue.

\subsection{Yaw}
\begin{figure*}
  \begin{minipage}[c]{.4\linewidth}
    \centering
    \includegraphics[width=0.9\linewidth, height=0.25\textheight]{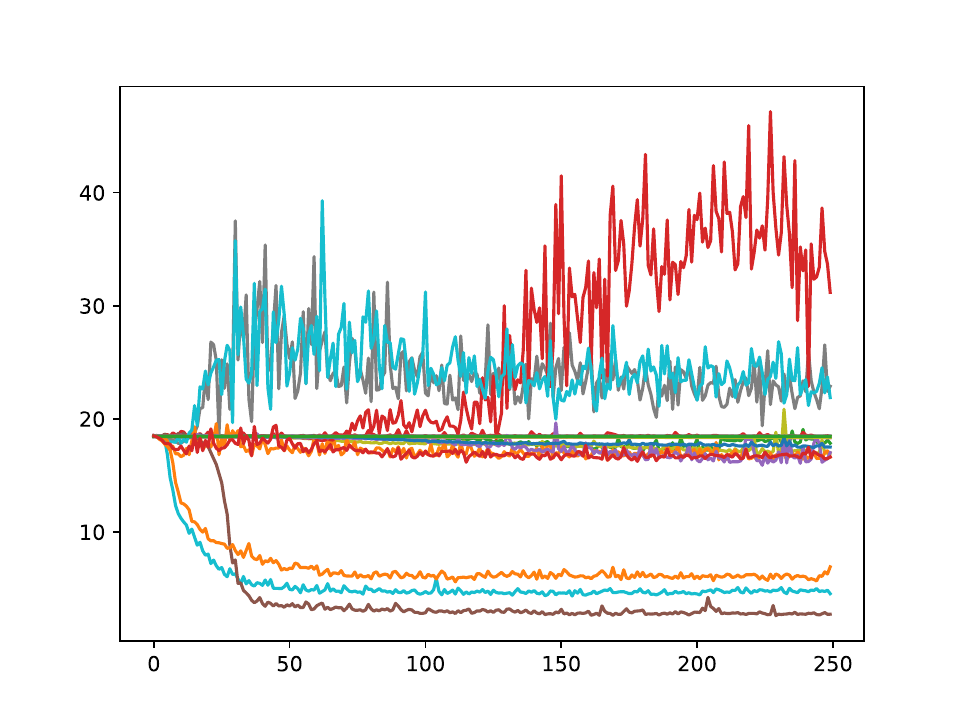}
    \captionof{figure}{Yaw Validation Loss (MSE deg/s)}
    \label{fig:yaw_graph}
  \end{minipage}\hfill
  \begin{minipage}[c]{.6\linewidth}
    \centering
    \begin{tabular}{|c|c|c|c|c|c|c|c|}
        \cline{3-8}
        \multicolumn{1}{c}{} && \multicolumn{2}{|c|}{Baseline} & \multicolumn{2}{|c|}{GRU}&\multicolumn{2}{|c|}{Transformer}\\
        \hline
        BS & LR & CNN & UNET & CNN & UNET & CNN & UNET \\
        \hline
        \multirow{2}{*}{1} & $10^{-3}$ & 18.439 & 18.439 & 18.446 & 17.840 & 18.439 & NaN \\
        \cline{2-8}
         & $10^{-5}$ & 22.873 & 18.325 & 4.539 & 16.961 & 16.612 & 18.437 \\
         \hline
        \multirow{2}{*}{2} & $10^{-3}$ & 31.181 & 18.438 & 2.726 & 17.005 & 18.439 & 18.439 \\
        \cline{2-8}
         & $10^{-5}$ & 21.877 & 18.351 & 6.908 & 17.502 & 16.624 & 18.439 \\
        \hline
    \end{tabular}
    \captionof{table}{Yaw (MSE deg/s)}
    \label{tab:yaw_results}
  \end{minipage}
\end{figure*}

The results of the 24 yaw models are shown in Table~\ref{tab:yaw_results}, and the validation loss across the epochs is displayed in Figure~\ref{fig:yaw_graph}.

Three of the models performed well for yaw, all of which were GRU and CNN. The ability of the GRU model to learn temporal relationships likely helped since a turn occurs over a couple of seconds. The CNN performed better than the autoencoder because the autoencoder is trained on speed, while yaw is only in this dataset. Since the yaw ranges between -40 and 40, an minimum MSE of 2.726 is promising. On the other hand, the car drives straight in the majority of the footage, so yaw would also benefit notably from more data.

\subsection{Lead Present}
\begin{figure*}
  \begin{minipage}[c]{.4\linewidth}
    \centering
    \includegraphics[width=0.9\linewidth]{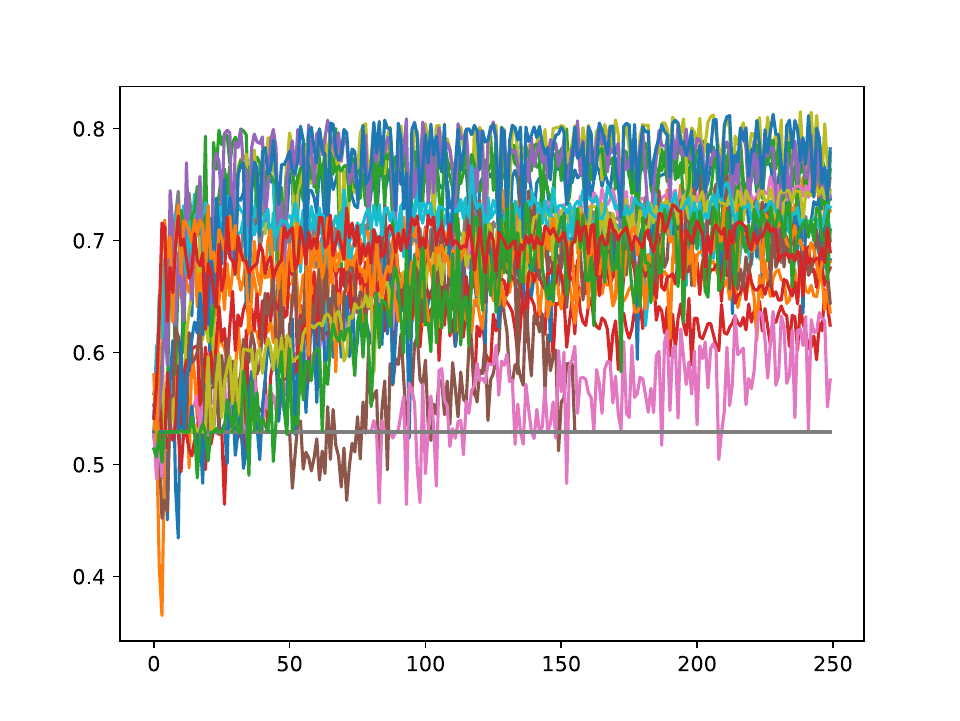}
    \captionof{figure}{Lead Car Present Validation Loss (accuracy)}
    \label{fig:lp_graph}
  \end{minipage}\hfill
  \begin{minipage}[c]{.6\linewidth}
    \centering
    \begin{tabular}{|c|c|c|c|c|c|c|c|}
        \cline{3-8}
        \multicolumn{1}{c}{} && \multicolumn{2}{|c|}{Baseline} & \multicolumn{2}{|c|}{GRU}&\multicolumn{2}{|c|}{Transformer}\\
        \hline
        BS & LR & CNN & UNET & CNN & UNET & CNN & UNET \\
        \hline
        \multirow{2}{*}{1} & $10^{-3}$ & 0.727 & 0.737 & 0.624 & 0.763 & 0.529 & 0.529 \\
        \cline{2-8}
         & $10^{-5}$ & 0.700 & 0.755 & 0.710 & 0.771 & 0.636 & 0.683 \\
         \hline
        \multirow{2}{*}{2} & $10^{-3}$ & 0.676 & 0.778 & 0.644 & 0.738 & 0.523 & 0.575 \\
        \cline{2-8}
         & $10^{-5}$ & 0.730 & 0.745 & 0.681 & 0.781 & 0.690 & 0.709 \\
        \hline
    \end{tabular}
    \captionof{table}{Lead Car Present (accuracy)}
    \label{tab:lp_results}
  \end{minipage}
\end{figure*}

The results of the 24 speed models are shown in Table~\ref{tab:lp_results}, and the validation loss across the epochs is displayed in Figure~\ref{fig:lp_graph}.

The best combinations of parameters had accuracies between 0.75 and 0.8. This is significantly higher than random chance, and it is also better than guessing one or the other because the data is split roughly equally between having a lead car and not having a lead car. The baseline and GRU models performed about the same, with the transformer performing worse. The autoencoder performed somewhat better than the CNN for this category, likely because the CNN is prone to overfit. Error may also be due to data collection. The cars are unable to detect other cars that are too far ahead, and may also detect non-car objects as lead cars if they behave similarly.
While these accuracies seem decent, they may have implications for lead distance and lead relative speed, because if the models already have trouble detecting if a car is there at all, there will be less accuracy when determining specific information about the cars.

\subsection{Lead Distance}
\begin{figure*}
  \begin{minipage}[c]{.4\linewidth}
    \centering
    \includegraphics[width=0.9\linewidth]{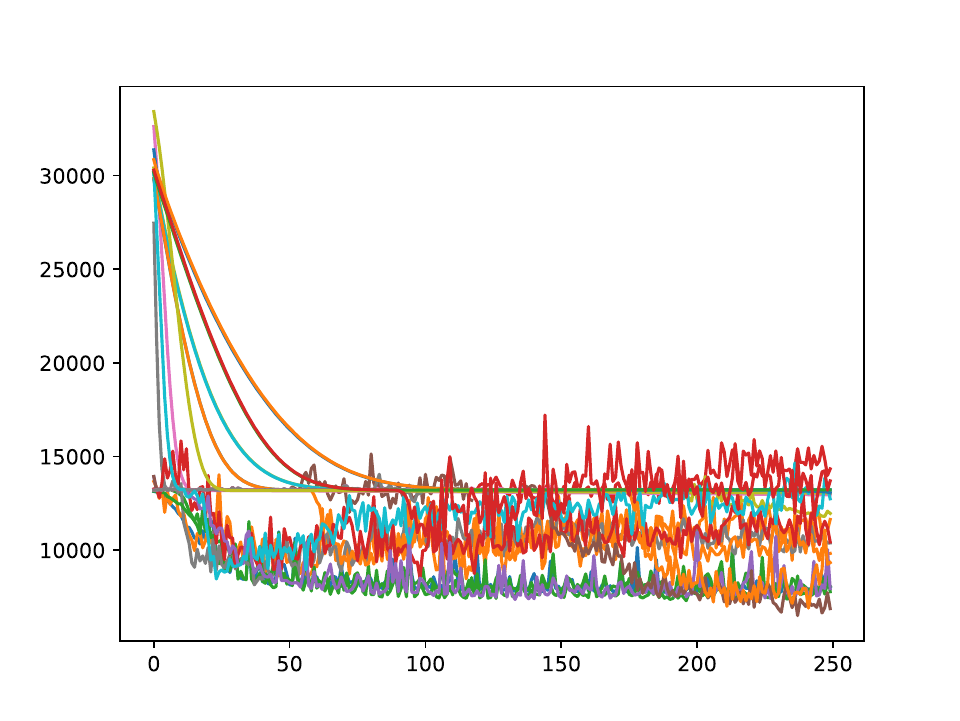}
    \captionof{figure}{Lead Distance Validation Loss (MSE m)}
    \label{fig:ld_graph}
  \end{minipage}\hfill
  \begin{minipage}[c]{.6\linewidth}
    \centering
    \begin{tabular}{|c|c|c|c|c|c|c|c|}
        \cline{3-8}
        \multicolumn{1}{c}{} && \multicolumn{2}{|c|}{Baseline} & \multicolumn{2}{|c|}{GRU}&\multicolumn{2}{|c|}{Transformer}\\
        \hline
        BS & LR & CNN & UNET & CNN & UNET & CNN & UNET \\
        \hline
        \multirow{2}{*}{1} & $10^{-3}$ & 11328 & 8100 & 14336 & 8064 & NaN & NaN \\
        \cline{2-8}
         & $10^{-5}$ & 10383 & 12950 & 13219 & 11960 & 9822 & 13219 \\
         \hline
        \multirow{2}{*}{2} & $10^{-3}$ & 10433 & 7779 & 6874 & 7945 & NaN & NaN \\
        \cline{2-8}
         & $10^{-5}$ & 114741 & 12042 & 9333 & 13062 & 13696 & 13218 \\
        \hline
    \end{tabular}
    \captionof{table}{Lead Distance (MSE m)}
    \label{tab:ld_results}
  \end{minipage}
\end{figure*}

The results of the 24 lead distance models are shown in Table~\ref{tab:ld_results}, and the validation loss across the epochs is displayed in Figure~\ref{fig:ld_graph}.

The models seemed to either fall into the MSE range of 6000-8000, or $>10000$. The models that fell into the lower range may have learned some information, but it is likely not accurate enough to be informative or trustworthy, since the MSE is large. Part of this is due to how the data is processed and encoded. Any time a lead car is not detected, the distance is recorded as 252. This interferes with the training both because this value does not properly reflect the distance of the car, and because the number is large enough to reflect poorly in the MSE.

\begin{figure*}
  \begin{minipage}[c]{.4\linewidth}
    \centering
    \includegraphics[width=0.9\linewidth, height=0.22\textheight]{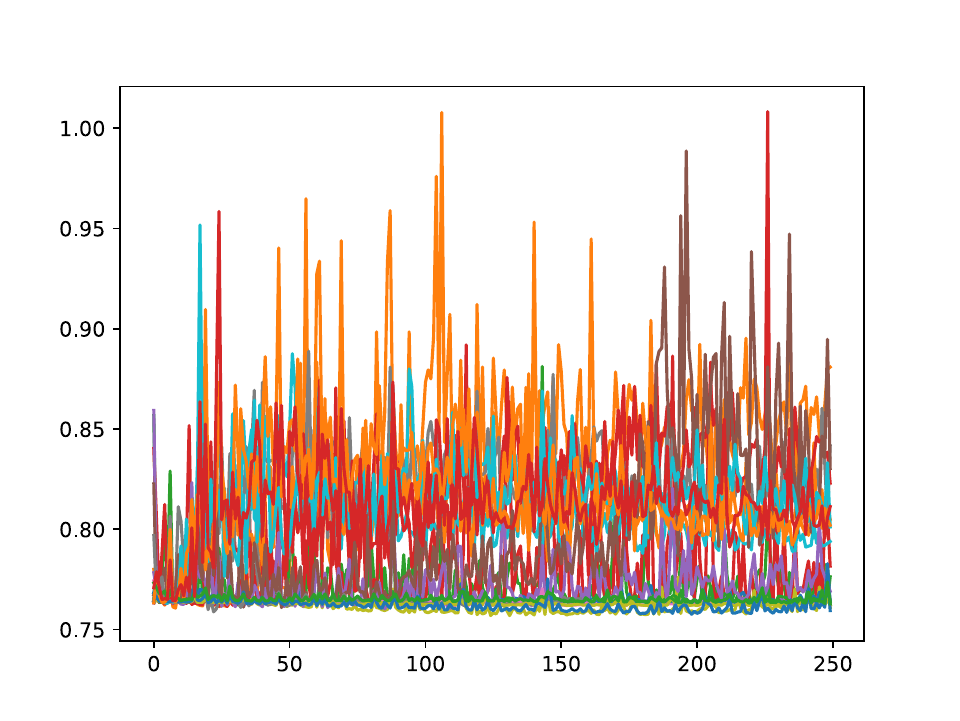}
    \captionof{figure}{Lead Relative Speed Validation Loss (MSE km/h)}
    \label{fig:ls_graph}
  \end{minipage}\hfill
  \begin{minipage}[c]{.6\linewidth}
    \centering
    \begin{tabular}{|c|c|c|c|c|c|c|c|}
        \cline{3-8}
        \multicolumn{1}{c}{} && \multicolumn{2}{|c|}{Baseline} & \multicolumn{2}{|c|}{GRU}&\multicolumn{2}{|c|}{Transformer}\\
        \hline
        BS & LR & CNN & UNET & CNN & UNET & CNN & UNET \\
        \hline
        \multirow{2}{*}{1} & $10^{-3}$ & 0.881 & 0.764 & 0.782 & 0.776 & NaN & 0.764 \\
        \cline{2-8}
         & $10^{-5}$ & 0.852 & 0.762 & 0.794 & 0.761 & 0.805 & 0.764 \\
         \hline
        \multirow{2}{*}{2} & $10^{-3}$ & 0.823 & 0.764 & 0.830 & 0.769 & NaN & NaN \\
        \cline{2-8}
         & $10^{-5}$ & 0.801 & 0.762 & 0.802 & 0.759 & 0.811 & 0.763 \\
        \hline
    \end{tabular}
    \captionof{table}{Lead Relative Speed (MSE km/h)}
    \label{tab:ls_results}
  \end{minipage}
\end{figure*}

\subsection{Lead Relative Speed}
The results of the 24 speed models are shown in Table~\ref{tab:ls_results}, and the validation loss across the epochs is displayed in Figure~\ref{fig:ls_graph}.

The relative speed of the lead car has many similar issues as the lead distance. Additionally, when a lead car is present, both cars are often driving at the same speed, so data where lead relative speed is of a significant magnitude is rare, and the models were unable to train well.

\section{Improvements}
Training these models on a larger dataset will significantly improve the validation loss of these models, as evidenced by the low MSE of the training performed on BDD100k. Additionally, possessing more data will likely mitigate the issue of different distributions between training and validation datasets, although better division of the datasets may also improve this.

For the issues caused by whether a lead car is present or not, the lead distance and lead relative speed could be trained on data where a lead car is present. This will remove the 252 issue from lead distance, and allow for less complex training overall. The models' ability to detect a lead car has an accuracy of 0.778 (which can be improved with a larger dataset), so in practice, the lead car information can be estimated after it is determined there is a lead car.

Unfortunately, removing videos without lead cars present throughout the video may result in significant loss of data. However, the experiment design is based roughly on the structure of the BDD100k dataset, and is not necessarily the optimal design for what we want to achieve. Since most events that happen in driving happen in a relatively short amount of time (turns, lane switches), it may be useful to reduce the 40 second chunks to around 10 seconds instead, and start the removal process from there.
Finally, it may be more informative and more accurate if lead relative speed is instead a classifier that determines whether the lead car is not moving, getting further, or getting closer. The exact speed of a car is not informative when it comes to events, and using a classifier may result in higher accuracy. 

\section{Conclusion and Future Work}
We explored the feasibility of training neural networks on dashcam footage to extract important driving-related information such as speed, yaw, lead car presence, lead car distance, and lead car relative speed. 
While our findings show potential, a major limitation is the relatively small size of our dataset, which impacted the ability of our models to train well. However, should additional data become available in the future the performance of these models should improve drastically. 
The results of this study can be used in extracting driving-related information for event detection, including turns, lane changes, and passing maneuvers, which is a useful tool in a wide variety of applications related to autonomous vehicles.

\section*{ACKNOWLEDGMENT}
This work is supported by the National Science Foundation through grant NSF-2135579, through a Research Experiences for Undergraduates supplement. 

\addtolength{\textheight}{-14cm}   

\bibliographystyle{IEEEtran}
\bibliography{sample-base}

@ARTICLE{bhadani2022strym,
  title={Strym: A python package for real-time can data logging, analysis and visualization to work with usb-can interface},
  author={Bhadani, Rahul and Bunting, Matt and Nice, Matthew and Tran, Ngoc Minh and Elmadani, Safwan and Work, Dan and Sprinkle, Jonathan},
  booktitle={2022 2nd Workshop on Data-Driven and Intelligent Cyber-Physical Systems for Smart Cities Workshop (DI-CPS)},
  pages={14--23},
  year={2022},
  organization={IEEE}
}

@ARTICLE{yu2020bdd100k,
  title={Bdd100k: A diverse driving dataset for heterogeneous multitask learning},
  author={Yu, Fisher and Chen, Haofeng and Wang, Xin and Xian, Wenqi and Chen, Yingying and Liu, Fangchen and Madhavan, Vashisht and Darrell, Trevor},
  booktitle={Proceedings of the IEEE/CVF conference on computer vision and pattern recognition},
  pages={2636--2645},
  year={2020}
}

@ARTICLE{attentionisallyouneed,
  author       = {Ashish Vaswani and
                  Noam Shazeer and
                  Niki Parmar and
                  Jakob Uszkoreit and
                  Llion Jones and
                  Aidan N. Gomez and
                  Lukasz Kaiser and
                  Illia Polosukhin},
  title        = {Attention Is All You Need},
  journal      = {CoRR},
  volume       = {abs/1706.03762},
  year         = {2017},
  url          = {http://arxiv.org/abs/1706.03762},
  eprinttype    = {arXiv},
  eprint       = {1706.03762},
  bibsource    = {dblp computer science bibliography, https://dblp.org}
}

@ARTICLE{unetautoencoder,
  author       = {Olaf Ronneberger and
                  Philipp Fischer and
                  Thomas Brox},
  title        = {U-Net: Convolutional Networks for Biomedical Image Segmentation},
  journal      = {CoRR},
  volume       = {abs/1505.04597},
  year         = {2015},
  url          = {http://arxiv.org/abs/1505.04597},
  eprinttype    = {arXiv},
  eprint       = {1505.04597},
  bibsource    = {dblp computer science bibliography, https://dblp.org}
}

@INPROCEEDINGS{KITTI,
  author={Geiger, Andreas and Lenz, Philip and Urtasun, Raquel},
  booktitle={2012 IEEE Conference on Computer Vision and Pattern Recognition}, 
  title={Are we ready for autonomous driving? The KITTI vision benchmark suite}, 
  year={2012},
  volume={},
  number={},
  pages={3354-3361},
  doi={10.1109/CVPR.2012.6248074}}

@ARTICLE{apollo,
  author={Huang, Xinyu and Wang, Peng and Cheng, Xinjing and Zhou, Dingfu and Geng, Qichuan and Yang, Ruigang},
  journal={IEEE Transactions on Pattern Analysis and Machine Intelligence}, 
  title={The ApolloScape Open Dataset for Autonomous Driving and Its Application}, 
  year={2020},
  volume={42},
  number={10},
  pages={2702-2719},
  doi={10.1109/TPAMI.2019.2926463}}

@INPROCEEDINGS{waymo,
  author={Sun, Pei and Kretzschmar, Henrik and Dotiwalla, Xerxes and Chouard, Aurélien and Patnaik, Vijaysai and Tsui, Paul and Guo, James and Zhou, Yin and Chai, Yuning and Caine, Benjamin and Vasudevan, Vijay and Han, Wei and Ngiam, Jiquan and Zhao, Hang and Timofeev, Aleksei and Ettinger, Scott and Krivokon, Maxim and Gao, Amy and Joshi, Aditya and Zhang, Yu and Shlens, Jonathon and Chen, Zhifeng and Anguelov, Dragomir},
  booktitle={2020 IEEE/CVF Conference on Computer Vision and Pattern Recognition (CVPR)}, 
  title={Scalability in Perception for Autonomous Driving: Waymo Open Dataset}, 
  year={2020},
  volume={},
  number={},
  pages={2443-2451},
  doi={10.1109/CVPR42600.2020.00252}}

@ARTICLE{9913431,
  author={Rafiq, Ghazala and Rafiq, Muhammad and On, Byung-Won and Sung, Mankyu and Choi, Gyu Sang},
  journal={IEEE Access}, 
  title={DeepRide: Dashcam Video Description Dataset for Autonomous Vehicle Location-Aware Trip Description}, 
  year={2022},
  volume={10},
  number={},
  pages={107361-107375},
  doi={10.1109/ACCESS.2022.3212745}}

@ARTICLE{9686618,
  author={Simoncini, Matteo and de Andrade, Douglas Coimbra and Taccari, Leonardo and Salti, Samuele and Kubin, Luca and Schoen, Fabio and Sambo, Francesco},
  journal={IEEE Transactions on Intelligent Transportation Systems}, 
  title={Unsafe Maneuver Classification From Dashcam Video and GPS/IMU Sensors Using Spatio-Temporal Attention Selector}, 
  year={2022},
  volume={23},
  number={9},
  pages={15605-15615},
  doi={10.1109/TITS.2022.3142672}}

@INPROCEEDINGS{8569952,
  author={Taccari, Leonardo and Sambo, Francesco and Bravi, Luca and Salti, Samuele and Sarti, Leonardo and Simoncini, Matteo and Lori, Alessandro},
  booktitle={2018 21st International Conference on Intelligent Transportation Systems (ITSC)}, 
  title={Classification of Crash and Near-Crash Events from Dashcam Videos and Telematics}, 
  year={2018},
  volume={},
  number={},
  pages={2460-2465},
  doi={10.1109/ITSC.2018.8569952}}

@inproceedings{10.1145/3383812.3383817,
author = {Chen, Jui-Chi and Lian, Zhen-You and Huang, Chung-Lin and Chuang, Cheng-Hung},
title = {Automatic Recognition of Driving Events Based on Dashcam Videos},
year = {2020},
isbn = {9781450377201},
publisher = {Association for Computing Machinery},
address = {New York, NY, USA},
url = {https://doi.org/10.1145/3383812.3383817},
doi = {10.1145/3383812.3383817},
booktitle = {Proceedings of the 2020 3rd International Conference on Image and Graphics Processing},
pages = {22–25},
numpages = {4},
location = {Singapore, Singapore},
series = {ICIGP '20}
}

@misc{caesar2020nuscenes,
      title={nuScenes: A multimodal dataset for autonomous driving}, 
      author={Holger Caesar and Varun Bankiti and Alex H. Lang and Sourabh Vora and Venice Erin Liong and Qiang Xu and Anush Krishnan and Yu Pan and Giancarlo Baldan and Oscar Beijbom},
      year={2020},
      eprint={1903.11027},
      archivePrefix={arXiv},
      primaryClass={cs.LG}
}

@misc{chung2014empirical,
      title={Empirical Evaluation of Gated Recurrent Neural Networks on Sequence Modeling}, 
      author={Junyoung Chung and Caglar Gulcehre and KyungHyun Cho and Yoshua Bengio},
      year={2014},
      eprint={1412.3555},
      archivePrefix={arXiv},
      primaryClass={cs.NE}
}

@ARTICLE{lstm,
  author={Hochreiter, Sepp and Schmidhuber, Jürgen},
  journal={Neural Computation}, 
  title={Long Short-Term Memory}, 
  year={1997},
  volume={9},
  number={8},
  pages={1735-1780},
  doi={10.1162/neco.1997.9.8.1735}}

\end{document}